\def\year{2019}\relax
\documentclass[letterpaper]{article} 
\usepackage{aaai19}  
\usepackage{times}  
\usepackage{helvet}  
\usepackage{courier}  
\usepackage{url}  
\usepackage{graphicx}  
\usepackage{subfigure}
\usepackage{multirow}
\usepackage{amsmath,amssymb}
\usepackage[linesnumbered,ruled]{algorithm2e}
\usepackage{bm}
\usepackage{color}
\usepackage{booktabs}

\def\C{\bm{C}} 
\def\c{\bm{c}} 
\def\D{\bm{D}} 
\def\d{\bm{d}} 
\def\E{\bm{E}} 
\def\e{\bm{e}} 
\def\h{\bm{h}} 
\def\i{\bm{i}} 
\def\o{\bm{o}} 
\def\p{\bm{p}} 
\def\s{\bm{s}} 
\def\u{\bm{u}} 
\def\v{\bm{v}} 
\def\W{\bm{W}} 
\def\w{\bm{w}} 
\def\x{\bm{x}} 
\def\y{\bm{y}} 
\def\model{Bi-STDDP } 
\def\setU{\mathbb{U}} 
\def\setP{\mathbb{P}}
\def\setR{\mathbb{R}}

\title{Modelling of Bi-directional Spatio-Temporal Dependence and Users' Dynamic Preferences for Missing POI Check-in Identification}
\author{
Dongbo Xi,\textsuperscript{\rm 1,2}
Fuzhen Zhuang,\textsuperscript{\rm 1,2,}\thanks{Corresponding author: Fuzhen Zhuang}
Yanchi Liu,\textsuperscript{\rm 3}
Jingjing Gu,\textsuperscript{\rm 4}
Hui Xiong,\textsuperscript{\rm 5}
Qing He\textsuperscript{\rm 1,2}\\
\textsuperscript{\rm 1}Key Lab of Intelligent Information Processing of Chinese Academy of Sciences (CAS),\\
Institute of Computing Technology, CAS, Beijing 100190, China\\
\textsuperscript{\rm 2}University of Chinese Academy of Sciences, Beijing 100049, China\\
\textsuperscript{\rm 3}Management Science \& Information Systems, Rutgers University, USA\\
\textsuperscript{\rm 4}Nanjing University of Aeronautics and Astronautics, Nanjing, China\\
\textsuperscript{\rm 5}Business Intelligence Lab, Baidu Inc., Beijing, China\\
\{xidongbo17s, zhuangfuzhen, heqing\}@ict.ac.cn,
yanchi.liu@rutgers.edu,
gujingjing@nuaa.edu.cn,
xionghui@gmail.com
}

\begin{document}
\setlength{\titlebox}{2.315in}
\maketitle
\begin{abstract}
Human mobility data accumulated from Point-of-Interest (POI) check-ins provides great opportunity for user behavior understanding. 
However, data quality issues (e.g., geolocation information missing, unreal check-ins, data sparsity) in real-life mobility data limit the effectiveness of existing POI-oriented studies, e.g., POI recommendation and location prediction, when applied to real applications.
To this end, in this paper, we develop a model, named Bi-STDDP, which can integrate bi-directional spatio-temporal dependence and users' dynamic preferences, to identify the missing POI check-in where a user has visited at a specific time.
Specifically, we first utilize bi-directional global spatial and local temporal information of POIs to capture the complex dependence relationships. 
Then, target temporal pattern in combination with user and POI information are fed into a multi-layer network to capture users' dynamic preferences. 
Moreover, the dynamic preferences are transformed into the same space as the dependence relationships to form the final model.
Finally, the proposed model is evaluated on three large-scale real-world datasets and the results demonstrate significant improvements of our model compared with state-of-the-art methods.
Also, it is worth noting that the proposed model can be naturally extended to address POI recommendation and location prediction tasks with competitive performances.
\end{abstract}

\section{Introduction}
Recent years have witnessed the rapid development and popularity of location-based social networks (LBSNs) and location-based services (LBS). These services have attracted many users to post their life experiences in the form of a check-in which contains a POI (a physical location, such as museum or restaurant), a timestamp, and sometimes comments. Data collected by LBS has been effectively leveraged for POI-oriented applications like POI recommendation and location prediction for potential users to improve user experiences and quality of services.

In this paper, we focus on missing POI check-in identification, which is to identify where a user has visited at a specific time in the past, while existing POI-oriented studies mainly focus on recommending or predicting a POI where a user may go in the future.
This missing POI check-in identification task is non-trivial due to the following reasons.
First, geolocation information missing, which causes trouble for user understanding, frequently occurs in raw real-life check-ins. For example, if a user visits a location without GPS signal, the geolocation information may be missing. 
And missing POI check-in identification helps to identify where a user has visited at given time. 
Second, there are many obvious unreal check-ins in raw real-life check-in data. For example, two check-in records of the same users are far apart in geographical distance, but the time interval is very short. The unreal data maybe recorded due to device failure or user cheating. 
Missing POI check-in identification can give a reference list and preliminarily filter the data.
Third, almost all kinds of POI recommendation and location prediction tasks suffer from data sparsity problem, which is much worse than other human activity data such as online purchase and browsing. 
And missing POI check-in identification is able to alleviate the sparsity problem for user understanding.
Moreover, the task can be used for social good, such as identifying traces of criminal activity or missing population analysis. 

Recently, efforts have been made to solve the POI recommendation and location prediction problems, which are most related to our problem. 
Factorizing Personalized Markov Chain (FPMC) \cite{rendle2010factorizing} is one widely used method for sequential prediction, and has been applied to embed the personalized Markov chains and the localized regions (FPMC-LR) \cite{cheng2013you} for next-basket recommendation. Personalized Ranking Metric Embedding (PRME) \cite{feng2015personalized} integrated sequence, preference and geographical influence to improve the recommendation performance. An unsupervised approach \cite{bao2012unsupervised} has also been adopted to model personalized contexts of mobile users.
Moreover, neural networks have been used in the field of POI recommendation and location prediction \cite{liu2016predicting,wang2017deep,yin2017spatial}. A method called Spatial Temporal Recurrent Neural Networks (ST-RNN) \cite{liu2016predicting} was proposed to model local temporal and spatial contexts in each layer with time-specific transition matrices for different time intervals and distance-specific transition matrices for different geographical distances. 
However, these methods are not designed for missing POI check-in identification.
They can not model global spatial information and spatio-temporal dependence relationships, which are very important in check-ins.
Besides, these methods utilize past check-in information for future prediction or recommendation from a single direction perspective, while the missing POI check-in identification task needs to utilize the check-in information before and after the given time, which naturally calls for a bi-directional solution.

Along this line, we propose a novel \textbf{Bi}-directional \textbf{S}patial and \textbf{T}emporal \textbf{D}ependence and users' \textbf{D}ynamic \textbf{P}references (Bi-STDDP) model for missing POI check-in identification.  
The bi-directional model can utilize more information than the above recommendation and prediction methods and is more suitable for the proposed task. 
Moreover, \model integrates local temporal information and global spatial information. Therefore, \model can well model not only the local context relationships but also the global ones. 
As we know, check-ins have complex spatio-temporal dependence relationships. For example, it is impossible that two check-in records of the same user are far apart in geographical distance, but the time interval is very short. 
This kind of complex spatio-temporal dependence can be modeled by \model and learned automatically by gradient descent performed on the whole model. 
In addition, target temporal pattern in combination with user and POI information can capture users' dynamic preferences. 
Finally, the spatio-temporal dependence together with users' dynamic preferences provide accurate identification on missing POI check-ins. 

The main contributions of this work are listed as follows:
\begin{itemize}
\item  The proposed model can address the non-trivial missing POI check-in identification task using bi-directional sequences to model complex global spatial and local temporal dependence relationships and users' dynamic preferences.
\item The traditional POI recommendation and location prediction tasks can be seen as a special case of the missing POI check-in identification task. And the proposed model can be easily extended to address POI recommendation and location prediction tasks by using only forward sequence information.
\item
Experimental results on real-world datasets show the proposed \model model obtains significant improvement compared with existing state-of-the-art approaches.
\end{itemize}

\section{Related Work}
The most related work to missing POI check-in identification is location prediction and POI recommendation. In this section, we present the related work in twofold: general methods on location prediction and POI recommendation, and neural network based methods.

\subsection{Location Prediction and POI Recommendation}
The key difference between location prediction and POI recommendation is that POI recommendation focuses on recommending new locations which do not exist in users' historical check-ins while location prediction can predict repeated locations. However recent work often does not make a clear distinction between them, we discuss them together here.

Different from traditional recommendation tasks (e.g., product recommendation, movie recommendation), location prediction and POI recommendation are to predict the next location utilizing spatial, temporal information of users' historical check-in sequences rather than user-item rating matrix. Therefore, the sequential information plays a crucial role in the successive check-in sequences. 
Users' movement constraint was taken into account in FPMC-LR (Factoring Personalized Markov Chains and Localized Regions) \cite{cheng2013you} via exploiting the personalized Markov chain in the check-in sequence. Personalized Ranking Metric Embedding (PRME) \cite{feng2015personalized} integrated geographical influence to improve the recommendation performance. Similarly, embedding method has also been introduced in Graph-based Embedding (GE) \cite{xie2016learning}. Besides, information such as temporal effects \cite{gao2013exploring}, spatial-aware \cite{yin2017spatial}, behavior patterns \cite{he2016inferring}, category-aware \cite{he2017category}, various contexts \cite{yang2017bridging} have been studied accordingly.

\subsection{Neural Network for Location Prediction and POI Recommendation}
Recently, neural network based methods not only have been successfully applied in sequential click prediction \cite{zhang2014sequential}, but also have been extended to location prediction and POI recommendation. An extended RNN called Spatial Temporal Recurrent Neural Networks (ST-RNN) \cite{liu2016predicting} modeled local temporal and spatial contexts with time-specific transition matrices and distance-specific transition matrices. Spatial-Aware Hierarchical Collaborative Deep Learning model (SH-CDL) \cite{yin2017spatial} utilized heterogeneous features and spatial-aware personal preferences. Joint Network and Trajectory Model (JNTM) \cite{yang2017neural} jointly modeled social networks and mobile trajectories via neural network-based approach. Deep Context-aware POI Recommendation (DCPR) \cite{wang2017deep} adopted CNN and RNN for POI and user characteristics respectively. Auxiliary meta-data information (e.g., textual description or category labels) was taken into consideration in NEXT \cite{zhang2017next}. Some LSTM-based approaches \cite{zhu2017next,zhao2018go} tried to capture short-term and long-term characteristics via specifically designed gates. User preference over POIs and context associated with users and POIs were predicted simultaneously in PACE (Preference And Context Embedding) \cite{yang2017bridging}. 

\section{Methodology}
In this section, we first formulate the problem of missing POI check-in identification, then we present the details of the proposed \model model, which integrates bi-directional spatio-temporal dependence and users' dynamic preferences.

\subsection{Problem Statement}
Let $\setU=\{u_1,u_2,...,u_N\}$ be a set of $N$ users and $\setP=\{p_1,p_2,...,p_M\}$ be a set of $M$ POIs. Each POI $p$ is associated with its coordinate $\{x_p,y_p\}$, and each user $u$ is associated with a list of check-ins $\C^u=\{p^u_{t_1},p^u_{t_2},...,p^u_{t_T},\}$, where $p^u_{t_i}$ means user $u$ visit POI $p$ at time $t_i$. Assume the $t$th check-in $p^u_{t_t}$ of user $u$ is missing, the task is to identify which POI the user $u$ visited at a specific time $t_t$ according to the forward sequence before $t_t$, $\{p^u_{t_1},p^u_{t_2},...,p^u_{t_{t-1}}\}$, and the backward sequence after $t_t$, $\{p^u_{t_{t+1}},p^u_{t_{t+2}},...,p^u_{t_{T}}\}$. 

\subsection{Bi-directional Spatial and Temporal Dependence}
Spatial and temporal information has very complex dependence relationships, here we first introduce these two kinds of information separately, global spatial information and local temporal information.
Global spatial information means we consider the relationship between the targeting POI and all the other POIs. 
While only local temporal information is considered since the missing check-in $p^u_{t_t}$ of user $u$ is more related to the POIs which user $u$ visited at a short temporal interval before and after time $t_t$.

We first introduce the extraction of the global spatial information. For POI check-ins $p^u_{t_{t-1}}$ and $p^u_{t_{t+1}}$, we define the global spatial vectors of POI check-ins $p^u_{t_{t-1}}$ and $p^u_{t_{t+1}}$ as follows:
\begin{eqnarray}
\s_{t-1}&=&\frac{\D_{p^u_{t_{t-1}}}}{\sigma(\D_{p^u_{t_{t-1}}})},\\
\s_{t+1}&=&\frac{\D_{p^u_{t_{t+1}}}}{\sigma(\D_{p^u_{t_{t+1}}})},
\end{eqnarray}
where $\sigma$ is the standard deviation, $\D\in\setR^{M\times M}$ is the geographical distance matrix of all candidate POIs, which contains the global spatial information, and $\D_{p^u_{t_{t-1}}}$ is the geographical distance vector between POI $p^u_{t_{t-1}}$ and all candidate POIs. The geographical distance vector can be normalized by dividing the standard deviation.

Besides, local temporal interval to the target time $t_t$ is considered as follows:
\begin{eqnarray}
i_{t-1}&=&t_t-t_{t-1},\\
i_{t+1}&=&t_{t+1}-t_t.
\end{eqnarray}

Due to the complex dependence relationships between spatial and temporal, for examples, it is impossible to go far away at a short time interval, but it is possible to go to a close POI at a great time interval. So we need take the global spatial and local temporal into overall consideration to model the complex dependence relationships, firstly, we transform the local temporal interval into the same space $\setR^M$ as the global spatial vectors: 
\begin{eqnarray}
\i_{t-1}&=&f(\w_{t-1}i_{t-1}),\\
\i_{t+1}&=&f(\w_{t+1}i_{t+1}),
\end{eqnarray}
where $\w_{t-1}, \w_{t+1}\in\setR^M$ and the activation function $f(x)$ is chosen as a $tanh$ function $f(x)=\frac{e^{x}-e^{-x}}{e^{x}+e^{-x}}$.

Finally, the bi-directional dependence relationships between global spatial and local temporal interval are encoded as follows:
\begin{eqnarray}
\d_{t-1}&=&\s_{t-1}\odot\i_{t-1}\label{d_t-1},\\
\d_{t+1}&=&\s_{t+1}\odot\i_{t+1}\label{d_t+1},
\end{eqnarray}
the dependence relationships are modeled by element-wise product operator $\odot$ and can be learned automatically by gradient descent performed on the whole model.  
\subsection{Users' Dynamic Preferences}
Users' check-in preferences change with time and the preference variance exists in several scales which is observed in \cite{gao2013exploring}. For example, users usually visit coffee house at afternoon and bars at night; users are usually around the office in weekdays while go shopping on weekends. Our model captures the specific temporal pattern in two scales: hours of a day, and different days of a week. We split a week into weekday and weekend, and a day into the following five sessions: the morning $[8:00,11:30)$, the noon $[11:30,14:00)$, the afternoon $[14:00,17:30)$, the night $[17:30,22:00)$ and the rest one. Thus, the target time $t_t$ is encoded to one $7$-dim pattern vector $\v_t$ by setting the corresponding bits to 1 and the rests to 0. For example, target time $11:30\ AM\ Aug\ 25\ 2018$ on a Saturday can be encoded into a pattern vector as follows (The first two bits correspond to the weekday and weekend, and the rest ones correspond to the five sessions):
\begin{equation}
\v_t=[0\ 1\ 0\ 1\ 0\ 0\ 0]^\top.
\end{equation}

Then, we capture user and POI information with embedding layers. The embedding layers can be seen as performing the latent factor modeling for POI popularity and user preference. It learns two matrices $\E_p$ and $\E_u$, each row of which represents a POI and a user, respectively. If we use one-hot encoded POI $\p^u_{t_{t-k}}$, $\p^u_{t_{t+k}}\in\setR^M$ and user $\u\in\setR^N$ as input vectors, the outputs of embedding layers can be expressed as
\begin{eqnarray}
\e(\p^u_{t_{t-k}})&=&\E_p^\top\p^u_{t_{t-k}},\\
\e(\p^u_{t_{t+k}})&=&\E_p^\top\p^u_{t_{t+k}},\\
\e(\u)&=&\E_u^\top\u,
\end{eqnarray}
where $1\leq k\leq w$, and $w$ is the window width, $\E_p \in \setR^{M\times d}$ and $\E_u \in \setR^{N\times d}$ denote the embedding matrices for POIs and users, $d$ is the dimension of embedding vectors.

Finally, we utilize multiple hidden layers of feed-forward neural networks to transform POI popularity, user preference and target temporal pattern into the same space and add them up to model users' dynamic preferences:
\begin{eqnarray}
\h(\e(\p^u_{t_{t-w}}))&=&f(\sum_{1\leq k\leq w}\W_{k_{-}}\e(\p^u_{t_{t-k}})),\\
\h(\e(\p^u_{t_{t+w}}))&=&f(\sum_{1\leq k\leq w}\W_{k_{+}}\e(\p^u_{t_{t+k}})),\\
\h(\e(\u))&=&f(\W_u\e(\u)),\\
\h(\v_t)&=&f(\W_t\v_t)\label{time_pattern},\\
\c^u_t&=&\h(\e(\p^u_{t_{t-w}}))+\h(\e(\p^u_{t_{t+w}}))\nonumber\\
    &&+\h(\e(\u))+\h(\v_t)\label{c_t},
\end{eqnarray}
where $\W_{k_{-}}, \W_{k_{+}},\W_u\in\setR^{h\times d}$ and $\W_t\in\setR^{h\times 7}$ are the parameters of hidden layers, $h$ is the number of hidden units and $f$ is $tanh$ activation function. And the $\c^u_t$ are the dynamic preferences of user $u$.

\subsection{The Final \model Model}

\begin{figure}[htbp]
\begin{center}
\includegraphics[width=.95\columnwidth]{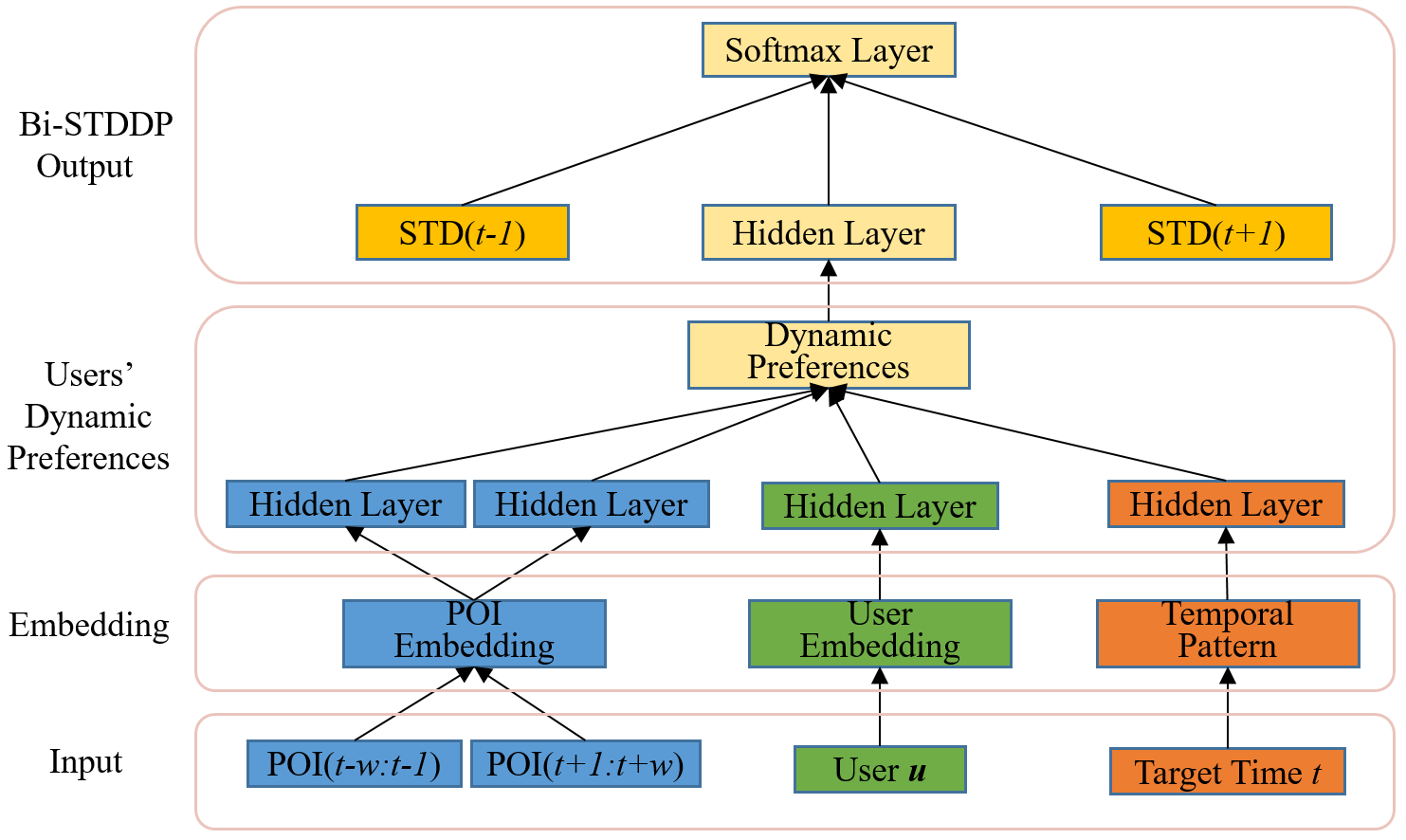}
\caption{The proposed \model model}
\label{fig:model}
\end{center}
\end{figure}

We present the final neural network architecture of \model in Figure \ref{fig:model}. 
The bi-directional check-in sequences, user and target time are fed into the embedding layers and users' dynamic preferences are captured by hidden layer.
The output of \model can be yielded via combining bi-directional spatial and temporal dependence and users' dynamic preferences in Equation (\ref{d_t-1}), (\ref{d_t+1}) and (\ref{c_t}) respectively. The prediction of where has user $u$ been at time $t_t$ can be computed as:
\begin{equation}
\o^u_t=softmax(\d_{t-1}+\d_{t+1}+\W_c\c^u_t),
\end{equation}
where $\W_c\in\setR^{M\times h}$ are the parameters to transform the users' dynamic preferences to the same space as the dependence relationships.
The $\o^u_t$ is a distribution which indicates different probability of all candidate POIs the user $u$ might visit at time $t_t$. 
And the POIs which k maximum probabilities corresponding to are the top-k identifications for the missing POI check-in. 

We need to minimize the \textit{cross entropy} of predicted distribution and the actual distribution: 
\begin{equation}
J(\theta)=-\frac{1}{S}\sum^S_{i=1}\sum^M_{j=1}y_{i,j}\log(o^u_{t,j}\vert \x_i,\theta),
\end{equation}
where $S$ is the number of samples, $M$ is the number of POIs, $\y_i\in\setR^M$ is the one-hot label of sample $\x_i$ and $\theta$ is the parameters set.

Training is done through stochastic gradient descent over shuffled mini-batches with the
Adam \cite{kinga2015method} update rule.

\section{Experiments}
In this section, we conduct experiments to evaluate the proposed \model against various baseline methods on three real-world datasets. In the next, we first introduce the datasets, implementation details, baseline methods and evaluation metrics, followed by our experimental results and discussions.

\subsection{Datasets}
Three real-world LBSN datasets, i.e., NYC, TKY, Gowalla, are used in the experimental study. The statistics of the three datasets are listed in Table \ref{tab:dataset}.

\begin{itemize}
\item \textbf{NYC}\footnote{https://sites.google.com/site/yangdingqi/home/foursquare-dataset\label{nyc}}  \cite{yang2015modeling} is a dataset from Foursquare, it includes long-term (about 10 months) check-in data in New York city collected from April 2012 to February 2013.
\item \textbf{TKY}\textsuperscript{\ref{nyc}} \cite{yang2015modeling} is a dataset similar to NYC except it is collected from Tokyo.
\item \textbf{Gowalla}\footnote{http://snap.stanford.edu/data/loc-gowalla.html} \cite{cho2011friendship} is a dataset collected from Gowalla, with the time span from February 2009 to October 2010.
\end{itemize}
\begin{table}[htbp]
  \centering
  \caption{Statistics of the three datasets.}
    \resizebox{.95\columnwidth}{!}{
    \begin{tabular}{ccccc}
    \toprule
    Dataset & \#user & \#POI & \#check\_in & Sparsity \\
    \midrule
    NYC   & 1,083  & 38,333 & 227,428 & 99.452\% \\
    TKY   & 2,293  & 61,858 & 573,703 & 99.596\% \\
    Gowalla & 107,092 & 1,280,969 & 6,442,892 & 99.995\% \\
    \bottomrule
    \end{tabular}%
    }
  \label{tab:dataset}%
\end{table}%

We eliminate users with fewer than 10 check-ins and POIs visited by fewer than 10 users in these three datasets. Then, we sort each user's check-in records by time, and take the first $80\%$ as the training set, the following $10\%$ as the validation set and the remaining $10\%$ as the test set.

\subsection{Implementation Details}
For all datasets we use: embedding dimension of 64, hidden units of 256, window width of 1 (temporal locality and a compromise between performance and efficiency), mini-batch size of 128 and learning rate of 0.001. All these values are chosen via a grid search on the NYC validation set. We initialize all parameters in the neural network from glorot uniform distributions \cite{glorot2010understanding}, and we do not perform any dataset-specific tuning except early stopping on validation sets.

\subsection{Baselines}
We compare the proposed method with counting-based methods (Forward, Backward, TOP1, TOP2), traditional POI recommendation algorithms (PRME, PRME-G), neural network based approaches (RNN, LSTM, GRU, STRNN, PACE). Some earlier methods likes PMF \cite{Salakhutdinov2007Probabilistic}, FPMC \cite{rendle2010factorizing}, FPMC-LR \cite{cheng2013you} have been proved to be not as good as PRME-G \cite{feng2015personalized,liu2016predicting,he2017category}, so we don't compare these methods. 

\begin{itemize}
\item \textbf{Forward}: The forward transition probability between POIs is taken as prediction for all users.
\item \textbf{Backward}: The backward transition probability between POIs is taken as prediction for all users.
\item \textbf{TOP1}: The most popular locations in the training set are selected as prediction for all users.
\item \textbf{TOP2}: The most popular locations in the training set are selected as prediction for each user.
\item \textbf{PRME}\cite{feng2015personalized}: User and POI are embedded into the same latent space to capture the user transition patterns.
\item \textbf{PRME-G}\cite{feng2015personalized}: It takes distance between destination location and recent visited ones into consideration on the basis of PRME.
\item \textbf{RNN}\cite{zhang2014sequential}: This is a neural network method which directly models the dependence on user's sequential behaviors into the click prediction process through the recurrent structure in RNN.
\item \textbf{LSTM}\cite{hochreiter1997long}: This is a special RNN model, which contains a memory cell and three multiplicative gates to learn long-term dependence.
\item \textbf{GRU}\cite{Cho2014Learning}: This is another special RNN model, which contains two gates and is simpler than LSTM.
\item \textbf{STRNN}\cite{liu2016predicting}: This is a RNN-based model for next POI recommendation. It incorporates both the time-specific transition matrices and distance-specific transition matrices within recurrent architecture.
\item \textbf{PACE}\cite{yang2017bridging}: This is a deep neural
architecture that jointly learns the embeddings of users and POIs
to predict both user preference over POIs and various context associated with users and POIs.
\end{itemize}

\subsection{Evaluation Metrics}
To evaluate the performance of our proposed \model and the baselines described above, we use three standard metrics following the existing work \cite{liu2016predicting}: \textbf{Recall@K}, \textbf{F1-score@K}, and Mean Average Precision (\textbf{MAP}). 
Note we don't use Precision@K since it is positively correlated with Recall@K and provides similar results in our settings. 
Recall@K is 1 if the POI visited appears in the top-K ranked list; otherwise is 0. The final Recall@K is the average value over all test ground truth instances. MAP is a global evaluation for ranking tasks, and it is usually used to evaluate the quality of the whole ranked lists. We report Recall@K and F1-score@K with K = 1, 5 and 10 in our experiments. The larger the value, the better the performance for all the evaluation metrics.

\begin{table*}[!th]
  \centering
  \caption{Evaluation of missing POI identification in terms of Recall@K, F1-score@K and MAP.}
    \begin{tabular}{ccccccccc}
    \toprule
          &       & {Recall@1} &{Recall@5} & {Recall@10} & {F1-score@1} &{F1-score@5} & {F1-score@10} & MAP \\
\cmidrule{1-9}    \multirow{12}[2]{*}{NYC} & Forward & 0.1065  & 0.2434  & 0.2801  & 0.1065  & 0.0811  & 0.0509  & 0.1647  \\
          & Backward & 0.1105  & 0.2399  & 0.2803  & 0.1105  & 0.0800  & 0.0510  & 0.1676  \\
          & TOP1  & 0.0045  & 0.0144  & 0.0211  & 0.0045  & 0.0048  & 0.0038  & 0.0112  \\
          & TOP2  & 0.1087  & 0.2899  & 0.3758  & 0.1087  & 0.0966  & 0.0683  & 0.1965  \\
          \cmidrule{2-9}
          & PRME  & 0.0987  & 0.2749  & 0.3568  & 0.0987  & 0.0916  & 0.0649  & 0.1796  \\
          & PRME-G & 0.1077  & 0.2836  & 0.3717  & 0.1077  & 0.0945  & 0.0676  & 0.1912  \\
          \cmidrule{2-9}
          & RNN   & 0.1264  & 0.3145  & 0.4012  & 0.1264  & 0.1048  & 0.0729  & 0.2131  \\
          & LSTM  & 0.1221  & 0.3136  & 0.4007  & 0.1221  & 0.1045  & 0.0729  & 0.2105  \\
          & GRU   & 0.1265  & 0.3177  & 0.4050  & 0.1265  & 0.1059  & 0.0736  & 0.2150  \\
          & STRNN & 0.1302  & 0.3225  & 0.4055  & 0.1302  & 0.1075  & 0.0737  & 0.2250  \\
          & PACE  & 0.1287  & 0.3226  & 0.4033  & 0.1287  & 0.1075  & 0.0733  & 0.2199  \\
          \cmidrule{2-9}
          & \model & \textbf{0.1743} & \textbf{0.3476} & \textbf{0.4176} & \textbf{0.1743} & \textbf{0.1159} & \textbf{0.0759} & \textbf{0.2533} \\
    \midrule
    \multirow{12}[2]{*}{TKY} & Forward & 0.1140  & 0.2618  & 0.3196  & 0.1140  & 0.0873  & 0.0581  & 0.1829  \\
          & Backward & 0.1302  & 0.2767  & 0.3300  & 0.1302  & 0.0922  & 0.0600  & 0.1979  \\
          & TOP1  & 0.0209  & 0.0798  & 0.1048  & 0.0209  & 0.0266  & 0.0191  & 0.0510  \\
          & TOP2  & 0.1370  & 0.3258  & 0.4002  & 0.1370  & 0.1086  & 0.0728  & 0.2257  \\
          \cmidrule{2-9}
          & PRME  & 0.0443  & 0.1149  & 0.1481  & 0.0443  & 0.0383  & 0.0269  & 0.0844  \\
          & PRME-G & 0.0870  & 0.2003  & 0.2557  & 0.0870  & 0.0668  & 0.0465  & 0.1436  \\
          \cmidrule{2-9}
          & RNN   & 0.1536  & 0.3590  & 0.4426  & 0.1536  & 0.1197  & 0.0805  & 0.2480  \\
          & LSTM  & 0.1468  & 0.3571  & 0.4450  & 0.1468  & 0.1190  & 0.0809  & 0.2431  \\
          & GRU   & 0.1514  & 0.3418  & 0.4457  & 0.1514  & 0.1139  & 0.0810  & 0.2478  \\
          & STRNN & 0.1612  & 0.3665  & 0.4492  & 0.1612  & 0.1222  & 0.0817  & 0.2543  \\
          & PACE  & 0.1608  & 0.3605  & 0.4469  & 0.1608  & 0.1202  & 0.0813  & 0.2518  \\
          \cmidrule{2-9}
          & \model & \textbf{0.2049} & \textbf{0.4107} & \textbf{0.4838} & \textbf{0.2049} & \textbf{0.1369} & \textbf{0.0880} & \textbf{0.2991} \\
    \midrule
    \multirow{12}[2]{*}{Gowalla} & Forward & 0.0809  & 0.1772  & 0.2239  & 0.0809  & 0.0591  & 0.0407  & 0.1268  \\
          & Backward & 0.0921  & 0.1880  & 0.2295  & 0.0921  & 0.0627  & 0.0417  & 0.1374  \\
          & TOP1  & 0.0029  & 0.0131  & 0.0232  & 0.0029  & 0.0044  & 0.0042  & 0.0105  \\
          & TOP2  & 0.0576  & 0.1331  & 0.1785  & 0.0576  & 0.0444  & 0.0325  & 0.0989  \\
          \cmidrule{2-9}
          & PRME  & 0.0228  & 0.0532  & 0.0703  & 0.0228  & 0.0177  & 0.0128  & 0.0394  \\
          & PRME-G & 0.0307  & 0.0678  & 0.0857  & 0.0307  & 0.0226  & 0.0156  & 0.0503  \\
          \cmidrule{2-9}
          & RNN   & 0.0630  & 0.1541  & 0.2053  & 0.0630  & 0.0514  & 0.0373  & 0.1120  \\
          & LSTM  & 0.0609  & 0.1489  & 0.1978  & 0.0609  & 0.0496  & 0.0360  & 0.1083  \\
          & GRU   & 0.0630  & 0.1555  & 0.2073  & 0.0630  & 0.0518  & 0.0377  & 0.1124  \\
          & STRNN & 0.0641  & 0.1601  & 0.2108  & 0.0641  & 0.0534  & 0.0383  & 0.1189  \\
          & PACE  & 0.0629  & 0.1587  & 0.2100  & 0.0629  & 0.0529  & 0.0382  & 0.1178  \\
          \cmidrule{2-9}
          & \model & \textbf{0.1019} & \textbf{0.2257} & \textbf{0.2881} & \textbf{0.1019} & \textbf{0.0752} & \textbf{0.0524} & \textbf{0.1653} \\
    \bottomrule
    \end{tabular}
  \label{tab:result}
\end{table*}%

\subsection{Performance Comparison}
The experimental results evaluated by Recall@K, F1-score@K, and MAP on NYC, TKY and Gowalla are presented in Table \ref{tab:result}. Since users' behavior patterns are regular and follow the long-tailed distribution,  
we see that counting-based methods Forward and Backward have acceptable performances on all three datasets. And the performances of Forward and Backward are even better than PRME and PRME-G. 
Similarly, counting-based personalized TOP2 also have an good performance on NYC and TKY, while the non-personalized TOP1 performs worst. PRME-G slightly improves the results comparing with PRME via incorporating distance information. RNN-based methods (RNN, LSTM, GRU) obtain similar performance improvement over PRME-G because of their sequence modeling capability. PACE predicts both user preference over POIs and various context associated with users and POIs to achieves further improvement over RNN-based methods. Another great improvement is brought by ST-RNN, and it is the best method among the baselines on three datasets. It incorporates both the time-specific transition matrices and distance-specific transition matrices within recurrent architecture. Moreover, \model outperforms the baseline methods over all evaluation metrics on all three datasets. On NYC dataset, the Recall@1, Recall@5, Recall@10 improvements comparing with best baseline ST-RNN are $33.87\%$, $7.78\%$, and $2.98\%$ respectively which indicates that the \model improves even more on higher ranking list, similar results can also be observed on TKY and Gowalla. These improvements indicate that our proposed \model can model the bi-directional dependence of global spatial and local temporal and capture users' dynamic preferences.

\subsection{Impact of Different Parts}
\begin{table}[!th]
  \centering
  \caption{Impact of forward and backward sequences on NYC dataset evaluated by Recall@K and MAP.}
  \resizebox{\linewidth}{!}{
    \begin{tabular}{ccccc}
    \toprule
        & Recall@1 &Recall@5 & Recall@10 & MAP \\
    \midrule
    F-STDDP & 0.1586  & 0.3378  & 0.4050  & 0.2405  \\
    B-STDDP & 0.1580  & 0.3372  & 0.4089  & 0.2405  \\
    \model & \textbf{0.1743} & \textbf{0.3476} & \textbf{0.4176} & \textbf{0.2533} \\
    \bottomrule
    \end{tabular}}
  \label{tab:bi-directional}%
\end{table}%

In this subsection, we firstly investigate the impact of forward and backward sequences. It is intuitive that bi-directional sequences can bring more useful information and performance improvement than a single sequence. As shown in Table \ref{tab:bi-directional}, \model improves the performance evaluated by Recall@K and MAP comparing with F-STDDP and B-STDDP which use forward and backward check-in sequences, respectively. Note that the performances of our method F-STDDP which only utilizes the forward sequence are still better than all baseline methods which contain POI recommendation and location prediction methods. It shows that the proposed \model model can be naturally applied to POI recommendation and location prediction tasks with competitive performance.

\begin{table}[!th]
  \centering
  \caption{Impact of temporal pattern and dependence relationships on NYC dataset evaluated by Recall@K and MAP.}
  \resizebox{\linewidth}{!}{
    \begin{tabular}{ccccc}
    \toprule
         &Recall@1&Recall@5 &Recall@10&MAP\\
    \midrule
    Bi-A &0.1435  & 0.3279  & 0.4068  & 0.2284  \\
    Bi-B &0.1650  & 0.3374  & 0.4128  & 0.2457  \\
    \model &\textbf{0.1743} & \textbf{0.3476} & \textbf{0.4176} & \textbf{0.2533} \\
    \bottomrule
    \end{tabular}}
  \label{tab:dependence}%
\end{table}%
Then, we study the impact of temporal pattern and dependence relationships. The \model is designed to model complex bi-directional spatio-temporal dependence relationships and capture users' dynamic preferences by target temporal pattern. By removing both dependence relationships and target temporal pattern in Equation (\ref{d_t-1}), (\ref{d_t+1}) and (\ref{time_pattern}) and only the dependence relationships in Equation (\ref{d_t-1}), (\ref{d_t+1}), we construct Bi-A and Bi-B models respectively for contrast experiment. Table \ref{tab:dependence} illustrates the performance of three models evaluated by Recall@K and MAP on NYC dataset. We see that Bi-B improves the results comparing with Bi-A by modeling the temporal pattern for users' dynamic preferences which indicates the significance of the temporal pattern. And \model achieves further improvement via taking the dependence relationships into consideration. So we can say that \model can capture dependence relationships between global spatial and local temporal and users' dynamic check-in preferences for improving the missing POI check-in identification performance.

\subsection{Impact of Parameters}
\begin{figure*}[!th]
\begin{center}
\subfigure[Impact of Embedding Dimension $d$ ($h=256$)]{ 
    \label{fig:embedding}
    \includegraphics[width=.95\columnwidth]{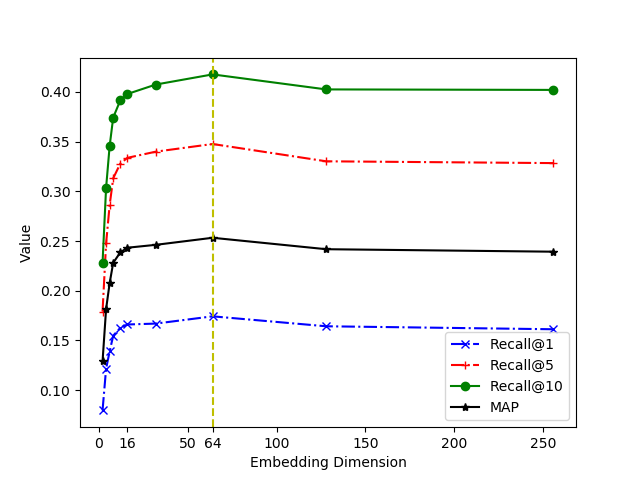} 
  } 
  \subfigure[Impact of Hidden Units $h$ ($d=64$)]{ 
    \label{fig:unit} 
    \includegraphics[width=.95\columnwidth]{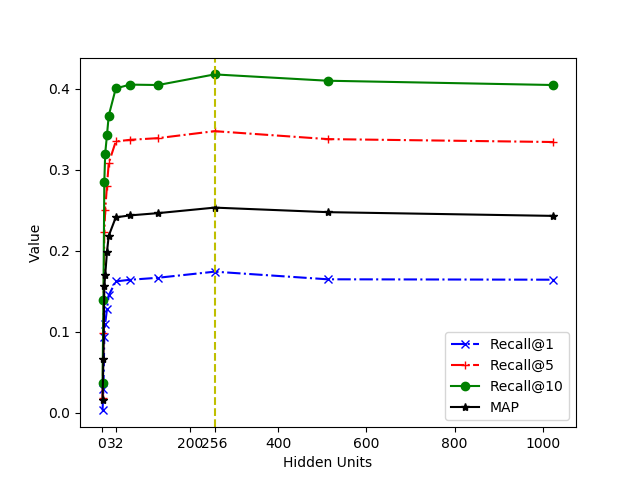} 
  } 
\caption{Performance of \model with varying embedding dimension and hidden units on NYC dataset evaluated by Recall@K and MAP.}
\label{fig:parameter}
\end{center}
\end{figure*}

Figure \ref{fig:embedding} and \ref{fig:unit} show the results under different settings of embedding dimension $d$ and hidden units $h$. We illustrate the Recall@K and MAP performance of \model on NYC test set. Note that the best parameters are selected by grid search on NYC validation set, while the impact of parameters is evaluated on NYC test set. Validation and test performances are similar on different parameters.

Firstly, considering embedding dimension, we vary the embedding dimension as $[2, 4, 6, 8, 12, 16, 32, 64, 128, 256]$. we can see that as the embedding dimension increases, the performance of the model is gradually improved, and when the embedding dimension $d$ larger than $16$, the performance becomes stabilized. The embedding dimension determines the complexity and capability of the model. Smaller embedding dimension may fit the data distribution insufficiently, especially if the numbers of POIs and users are large. While larger embedding dimension increases the complexity of the model and requires more computational cost, a proper embedding dimension can help achieve the best performance. Making a compromise between performance and efficiency, we finally select $d=64$ as the embedding dimension.

Similar results can also be observed with varying hidden units. We vary the number of hidden units as $[2,4,6,8,12,16,32,64,128,256,512,1024]$ to evaluate the performance on NYC dataset of our proposed \model model. As shown in Figure \ref{fig:unit}, the performance of the model becomes stabilized when the number of hidden units is larger than $32$. The hidden layer which is closer to the output requires larger feature dimension than embedding layer which is closer to the input. So $h=256$ is a proper number of hidden units.
Finally, we can get the best parameters which $d=64$ and $h=256$ respectively.
Even without tuned parameters, \model still outperforms other baseline methods according to Table \ref{tab:result} and Figure \ref{fig:parameter}. Further, we can say that the performance of \model stays stable in a large range of parameters and is not very sensitive to both embedding dimension and hidden units.
\begin{figure}[!htb]
\begin{center}
\includegraphics[width=.95\columnwidth]{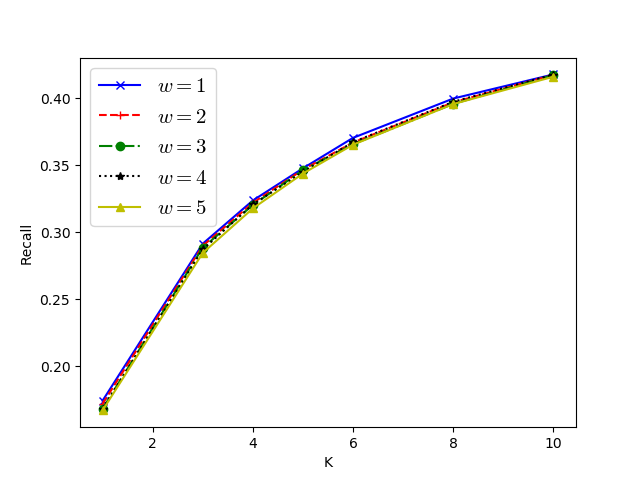}
\caption{Recall@K performance of \model on NYC dataset with varying window width $w$.}
\label{fig:window}
\end{center}%
\end{figure}%

We also vary the window width $w$ of our proposed model on NYC dataset to investigate the impact of the sequence length. Figure \ref{fig:window} shows the detailed Recall@K performance. There is no obvious difference between these results of different window width. It makes sense that the missing POI check-in $p^u_{t_t}$ of user $u$ is more related to the POIs which user $u$ visited at a short temporal interval before and after time $t_t$, and user preferences and periodic information have been modeled by target temporal pattern in combination with user and POI information which can learn automatically by gradient descent performed on the whole model of all training data, so larger window width brings little useful additional information and may even hurt the model performance due to introducing noise information. Therefore, making a balance between performance and efficiency, we select $w=1$ as the window width.

\section{Conclusion}
In this paper, we focused on the missing POI check-in identification task to identify where a user has visited at a specific time in the past, which is different from existing POI-oriented tasks to recommend or predict a POI where a user may go in the future. To address this task, we proposed a novel neural network model called Bi-STDDP. Specifically, in our model the bi-directional global spatial and local temporal information are combined together to capture complex dependence relationships. Also, target temporal pattern in combination with user and POI information are integrated to capture users' dynamic preferences. Finally, extensive experimental results on three large-scale real-world datasets demonstrated the substantial performance improvement of our proposed model over various kinds of state-of-the-art methods.

\section{Acknowledgments}
The research work is supported by the National Key Research and Development Program of China under Grant No. 2018YFB1004300, the National Natural Science Foundation of China under Grant No. 61773361, 61473273, 91546122, Guangdong provincial science and technology plan projects under Grant No. 2015B010109005, the Project of Youth Innovation Promotion Association CAS under Grant No. 2017146.

\bibliographystyle{aaai}
\end{document}